%% file: acl_latex.tex
\pdfoutput=1
\documentclass[11pt]{article}

\usepackage[]{acl}

\usepackage{times}
\usepackage{latexsym}

\usepackage[T1]{fontenc}
\usepackage[utf8]{inputenc}

\usepackage{microtype}

\usepackage[normalem]{ulem}

\title{Autism Detection in Speech -- A Survey}

\author{Nadine Probol \\
        University of Applied Sciences \\
        Darmstadt \\
        \texttt{nadine.probol@h-da.de } \\\And
  Margot Mieskes \\
        University of Applied Sciences \\
        Darmstadt \\
        \texttt{margot.mieskes@h-da.de }}

\begin{document}
\maketitle
\begin{abstract}
There has been a range of studies of how autism is displayed in voice, speech, and language.
We analyse studies from the biomedical, as well as the psychological domain, but also from the NLP domain in order to find linguistic, prosodic and acoustic cues that could indicate autism.
Our survey looks at all three domains.
We define autism and which comorbidities might influence the correct detection of the disorder.
We especially look at observations such as verbal and semantic fluency, prosodic features, but also disfluencies and speaking rate.
We also show word-based approaches and describe machine learning and transformer-based approaches both on the audio data as well as the transcripts. 
Lastly, we conclude, while there already is a lot of research, female patients seem to be severely under-researched. 
Also, most NLP research focuses on traditional machine learning methods instead of transformers which could be beneficial in this context. 
Additionally, we were unable to find research combining both features from audio and transcripts.% \textcolor{blue}{as indicators for autism }.
\end{abstract}

\input{content}

\bibliography{bibliography}
\bibliographystyle{acl_natbib}

\end{document}

%% file: content.tex
\section{Introduction}

With an increase in people with Autism Spectrum Disorder (ASD), the need for supporting the detection has increased as well.
Therefore, we look into research results from psychology, biomedicine,  as well as Natural Language Processing (NLP) to gain an insight how the three fields can support each other.

\textbf{Statistics}

The number of people with ASD varies depending on the source.
The German Federal Association for the Promotion of People with Autism (Bundesverband zur Förderung von Menschen mit Autismus) for example puts the frequency of all forms of autism spectrum disorders at 6-7 per 1,000, of which 1-3 per 1,000 are Asperger's autistics.\footnote{\url{https://www.autismus.de/was-ist-autismus.html}} 
The numbers are even higher (1 in 36 children) according to the Centers for Disease Control and Prevention (CDC) and have been increasing for years.\footnote{\url{https://www.cdc.gov/ncbddd/autism/data.html}}
Out of these people, about 25\% - 30\% are nonverbal or minimal-verbal \cite{posar2021}, though there are no concrete numbers.

\textbf{Wording and level of intelligence}

People with Asperger's syndrome (AS) tend to have an average to above average speech.
Their language often is very formal, direct and does not try to attempt to not offend others. \citet{hosseini2020} report "Lack of skills that are required to use language in context successfully (i.e., impaired pragmatic language) may cause ASD-AS subjects to communicate very formalized, direct, and without attempting to avoid offending others. This weakness can cause problems in the workplace, particularly in certain occupations such as work positions that need teamwork".

People with AS generally have a higher verbal IQ than performance IQ, however, they have an overall average to above average general IQ \cite{hosseini2020}.
This also means that they are often not diagnosed until they are adults, as their intelligence allows them to "mask" their deficits in communication and social interaction.
As they get older, it becomes more difficult to maintain this "masking" as the social environment becomes more complicated, so they are eventually diagnosed \cite{hosseini2020}.

\textbf{Sex differences}

Overall, the number of autistic people with a higher IQ is increasing \cite{baio2018}.
Although girls on the spectrum are more likely to display a lower IQ than boys \cite{zeidan2022}, this statistic takes into account all forms of autism and does not focus on AS or a comparable group withing the autism spectrum.

In general, women are diagnosed about seven to eleven years later than men \cite{breddemann2023}. The authors themselves even call this a "strong gender bias".

According to the CDC, autism is four times more common in boys and men than in girls and women.\footnote{\url{https://www.cdc.gov/ncbddd/autism/data.html}}
Similar numbers (4.5 times higher in boys than in girls) were obtained by \citet{christensen2016}.
However, this discrepancy has been decreasing for years, which is why there are voices that say that this difference is mainly due to the fact that girls and women are often not recognised.\footnote{\url{https://icd.who.int/browse11/l-m/en\#/http://id.who.int/icd/entity/437815624}, accessed August 22, 2023}
A reason for this discrepancy might be due to females on the spectrum displaying "fewer restricted, repetitive interests and behaviours" than their male counterparts.\footnote{\url{https://icd.who.int/browse11/l-m/en\#/http://id.who.int/icd/entity/437815624}, accessed August 22, 2023}

\textbf{Age}

When examining the data, taking into account the age of the participants is important. 
Generally, it is said the more data the better, especially with respect to machine learning methods, however, this does not apply to autism detection in speech.
As there are great differences in how the disorder presents itself in speech, it is important to differentiate according to age, especially when looking at children.
The best results in general can be achieved by using data from adults \cite{hauser2019} although a lot of studies have been done on data from children (see also Table~\ref{tab:mlapproaches}).

\vspace{0.5cm}

The research questions behind our work are:
One, what are gaps in the currently available research landscape?
Second, what are potential strategies to identify ASD in speech, based on research from various domains analysed here?

In the following, we take a look at the research from the biomedical and psychological field (Section~\ref{sec:biomedpsy}).
We define autism (Section~\ref{ssec:definition}) and describe comorbidities (Section~\ref{ssec:comorbidities}).
Then, we take a closer look at the verbal fluency of autistic patients (Section~\ref{ssec:bioverbalfluency}).
We describe prosodic approaches to examine autism in individuals (Section~\ref{ssec:bioprosodic}) with a special focus on the speaking rate (Section~\ref{ssec:speak_med}).
The second part of this work focuses on NLP research (Section~\ref{sec:nlp}) with a focus on research on prosodic features in Section~\ref{ssec:nlpprosodic}.
We take a closer look at the semantic fluency (Section~\ref{sssec:nlpsemantic}), the production of disfluencies (Section~\ref{sssec:nlpum}) and the speaking rate (Section~\ref{sssec:nlpspeakingrate}).

In Section~\ref{ssec:nlpml}, we describe machine learning approaches.
We look at approaches based on audio data (Section~\ref{sssec:nlpaudio}) separately from approaches based on transcripts (Section~\ref{sssec:nlptranscript}).
Then, we take a look at transformer-based approaches (Section~\ref{ssec:nlptransformer}).
We conclude our findings in Section~\ref{sec:conclusion} and present answers to our research questions.

\section{Bio-/Med-/Psych}
\label{sec:biomedpsy}

In order to find identifiers for ASD in speech, it is important to take a look at the medical descriptions and findings.

\subsection{Definition of Autism}
\label{ssec:definition}

The first description of Autism has been made by \citet{kanner1943}, followed by \citet{asperger1944}.
While \citet{kanner1943} describes autistic children as individuals who tend to avoid interacting with other people and having difficulties in learning the language (some even staying mute), \citet{asperger1944} describes his patients as having developed language at an early age, though not reacting to affective or emotional language at all.
However, the definition of autism has since changed.
Nowadays, there are mainly DSM-IV (Diagnostic and Statistical Manual of Mental Disorders IV) and ICD-10\footnote{\url{https://icd.who.int/browse10/2019/en}, accessed August 22, 2023} (International Statistical Classification of Diseases and Related Health Problems 10) as well as the newer versions DSM-V and ICD-11\footnote{\url{https://icd.who.int/browse11/l-m/en\#/http://id.who.int/icd/entity/437815624}, accessed August 22, 2023} used to define autism.
Whereas the DSM is a classification tool just for the US, ICD is published by the WHO\footnote{\url{https://www.who.int/}, accessed, August 22, 2023}.
Therefore, we focus on the definitions of the ICD-10 and ICD-11.

In the ICD-10 classification, autism is still differentiated into Childhood autism (which includes Kanner syndrome), atypical autism, and Asperger syndrome.\footnote{\url{https://icd.who.int/browse10/2019/en\#/F84.5}, accessed August 22, 2023}
The newer ICD-11, does not differentiate into these three types anymore but accumulates them under autistic spectrum disorder (ASD).\footnote{\url{https://icd.who.int/browse11/l-m/en\#/http://id.who.int/icd/entity/437815624}, accessed August 22, 2023}

The ICD-10\footnote{\url{https://icd.who.int/browse10/2019/en\#/F84.5}, accessed October 5, 2023} defines Asperger's Syndrome (AS) as "A disorder of uncertain nosological validity, characterized by the same type of qualitative abnormalities of reciprocal social interaction that typify autism, together with a restricted, stereotyped, repetitive repertoire of interests and activities."
It highlights that there is no general delay in languages as well as cognitive development.
However, ICD-10 sees AS as "often associated with marked clumsiness", which along with the other abnormalities tends to stay into adolescence and adult life.

The definition in ICD-11 is much longer and includes aspects which may or may not be included.
It describes ASD as "characterised by persistent deficits in the ability to initiate and to sustain reciprocal social interaction and social communication, and by a range of restricted, repetitive, and inflexible patterns of behaviour, interests or activities that are clearly atypical or excessive for the individual’s age and sociocultural context."\footnote{\url{https://icd.who.int/browse11/l-m/en\#/http://id.who.int/icd/entity/437815624}, accessed October 5, 2023}
The ICD-11 classification describes the disorder to occur in early childhood, however, it might fully manifest later, when the "social demands exceed the limited capacities", which impairs not only social life (including family) but can negatively affect educational and occupational life as well.

This is important to note, as a lot of research has been performed based on the definitions from before ICD-11. Additionally, some researchers still use these old definitions.
In our survey, we focus on research on Asperger Syndrom (AS) as well as high-functioning individuals with ASD in order to address the same group.

\subsection{Comorbidities}
\label{ssec:comorbidities}

Studies show 50\% to 70\% of ASD individuals are also diagnosed with attention deficit hyperactivity disorder (ADHD) \cite{hours2022}.

Other common comorbidities are Obsessive Compulsive Disorder (OCD) or Bipolar Disorder (BD) \cite{duda2016}.

Considering these numbers, it is hard to get data of individuals with just ASD and no co-occurring diagnoses.
As there is already very little data available, it is unlikely, to find enough individuals solely with ASD, but research is often done with individuals with ASD and at least one comorbidity.

\subsection{Verbal fluency}
\label{ssec:bioverbalfluency}

\citet{turner1999} studied so called High Functioning Autism (HFA) individuals, high functioning individuals without autism, learning disabled individuals, and learning disabled autistic individuals with respect to their verbal fluency.
For this, he asked the participants to produce as many words starting with the letters F, A, and S within 60 seconds as possible.
The same was done with categories, as the participants were asked to name as many words as possible of the categories animals, foods, and countries in 60 seconds.
The results lead \citet{turner1999} to the conclusion that verbal fluency correlates with executive function and therefore is linked to autism.

\citet{spek2009} conducted a study on the semantic and phonemic fluency.
To do so, the authors recruited participants aged 18 to 60 years, including 31 AS (29 male and 2 female), 31 individuals (28 male, 3 female) with HFA,\footnote{The ICD-10 only differentiates into childhood autism, atypical autism, and Asperger autism. It does not have a separate code for HFA as it did not differentiate based on IQ. However, the term HFA is sometimes used for autistic individuals with normal to high intelligence levels. The ICD-11 does not differentiate at all and subsumes all autistic individuals under ASD. Therefore, there is also no official code for HFA as it does not differentiate based on IQ as well.} and 30 so called neurotypical (NT) individuals (28 male, 2 female).
The authors found, that individuals with Asperger's syndrome have a similar fluency to neurotypical individuals.
This lead the authors to the hypothesis, that deficits in executive functioning reduces and even largely disappear when growing up.

Children and adolescents with ASD seem to use clustering as an efficient strategy to generate an equal number of words \cite{begeer2014}, which is also reported by \citet{turner1999}.
Clustering describes the strategy to find words, which are related to each other (e.g. farm animals; \citealt{turner1999,begeer2014}.

Though individuals with Asperger's Syndrome tend to display a higher verbal IQ than performance IQ \cite{hosseini2020}, their language contains some conspicuous features.
%\textcolor{red}{Average in what way???}
They tend to speak very formal and direct and do not even attempt to try to not offend other people \cite{hosseini2020}.

Even though research shows that there is no difference in the amount of correctly answered semantic fluency tasks in children and adolescents with ASD and NT individuals, a difference in strategy can be found \cite{dunn1996, begeer2014}.
This change of strategy might be visible in less prototypical answers \cite{dunn1996}.
\citet{begeer2014} found, ASD individuals have fewer switches, though in comparison to NT individuals, they produce slightly larger clusters.
For this, the authors examined 26 children and adolescents on the spectrum (23 male, 3 female) and 26 NT ones (22 male, 4 female).
The authors link these findings to different behaviour with regards to subcategories.
While NTs tend to switch in between subcategories more often, ASD individuals retrieve more words from just one subcategory.
%\textcolor{red}{Baron-Cohen et al., 2003} suspects this behaviour of ASD individuals to be connected to the preference of closed systems.
A reason for this might be special interests in some topic, which may lead to larger clusters in specific topics in ASD children than NT.
\citet{begeer2014} therefore hypothesise that stereotypical behaviour may not exclusively be an impairing feature but an asset.
Being able to get large amounts of information from a limited source, indicates that this allows individuals to compensate for other aspects.
This leads the authors to the conclusion that children and adolescents use clustering as a strategy to generate a comparable amount of words as NTs.
These findings contradict \citet{turner1999}, who observed that ASDs produce fewer words per cluster (based on 19 male, 3 female individuals).

\citet{begeer2014} also hypothesise that mature ASD participants overcome their limitations in verbal fluency as they reach similar amounts of words in these tasks due to their clustering strategy as NT participants.

\vspace{0.5cm}

\citet{asperger1944} described that individuals with AS display narrow and pedantic interests.
With respect to speech, the author described children to have a particularly creative relationship with language to explain their experiences and observations in a linguistic form.
He observed that children with AS use uncommon words, which one would not associate with the environment, the child grows up in.
Additionally, the children form completely new words or transform already existing words in order to fit their needs.
An example of this behaviour is a German speaking child saying "mündlich kann ich das nicht, aber köpflich" which can be translated to "I'm not able to do it verbally but headly".
"Headly" in this example means "doing something with the head" as opposed to doing it with words.
These words can be extremely fitting in some occasions, while being absolutely absurd in other ones.\footnote{Example was not given.}
\citet{luyster2022} point out, that this description is very important, as this form of language generation is associated with higher concurrent structural language skills.

Whereas \citet{asperger1944} only describes the interests of AS individuals as pedantic in general, some researchers describe only the speech of ASD individuals as "pedantic speech" (\citealp{luyster2022, neihart2000}; \citealp{wing1981} as in \citealp{ghaziuddin1996pedantic}).
\citet{neihart2000} focuses this definition mainly on gifted children with ASD.
There are different definitions of pedantic speech.
\citet{wing1981} as in \cite{ghaziuddin1996pedantic} describes it to be lengthy and "having a bookish quality."
According to \citet{wing1981} as in \cite{ghaziuddin1996pedantic}, AS individuals tend to use complicated and uncommon words.
To other people, this may seem like people with AS copy the speech of other people in an inappropriate way.
Individuals with AS have a much more pedantic speech than ones with High Functioning Autism (HFA) \cite{ghaziuddin1996pedantic}, though the Verbal IQ is higher in AS individuals.
Later, \citet{burgoine1983} list pedantic speech to be one of the major clinical features of ASD.

\subsection{Prosodic indicators}
\label{ssec:bioprosodic}

\citet{bone2012} studied prosodic features in children with ASD (22 male, 6 female).
The authors find the most important features are related to  monotone speech, variable volume and atypical voice quality.
The authors examined English and Spanish speaking children.
While negative average pitch slope at the end of turns is generally associated with statements, the authors observe that a lower average pitch slope also suggests a higher atypicality.
In general, this feature was only ever positive for children with the least atypical ratings.

These findings are supported by \citet{vogindroukas2022}, who describe acoustic studies, which show a greater intonational range in autistic individuals than in NTs.
Also, they find differences in prosodic phrasing as well as stress with respect to durational cues.

A study by \citet{plank2023} (ASD: 17 male, 18 female; NT: 21 male, 33 female) concluded, that autistic individuals have a lower pitch variance than non-autistic ones.
This study is especially interesting, as it is one of very few, which includes a balanced male-to-female ratio in its participants.% \textcolor{blue}{, especially for ASD}.

\subsubsection{Speaking rate}
\label{ssec:speak_med}

\citet{bone2012} found a correlation between speaking rate and being atypical.
The slower the speaking rate of a child, the more likely the child was evaluated as atypical as opposed to neurotypical.
The authors observed the "sixth and final correlated children’s prosody feature is the 90\% quantile syllabic speaking rate of nonturn-end words. This feature can be considered a robust measure of maximum speaking rate. A maxima was desired because it may indicate maximal ability, and other considered features capture rate variability."

This aligns with findings by \citet{vogindroukas2022}, who looked at language profiles of individuals with ASD.
The authors also describe studies to have confirmed that the speaking rate of ASD individuals is generally slower.

\section{NLP Research}
\label{sec:nlp}

\begin{table*}[htbp]
  \centering
  \begin{small}
  \begin{tabular}{|p{2cm}|p{1.5cm}|p{1.4cm}|p{2.2cm}|p{1.4cm}|p{1.5cm}|p{3cm}|}
  \hline
    Authors & Methods & Support \newline all(f) \newline NT/ASD  & Data volume & Data type & Age & Results \\
    \hline
    \hline
    \citet{bone2012} & Prosodic features & 28(6) \newline 6/22 & up to 5min per child (µ=264s, min=101s) & audio & mean 9.8 years & voice descriptions such as ‘breathy’, ’hoarse’, and ‘nasal’ are common in ASD children \\
    \hline
    \citet{parish2016HLT} & Prosodic features \newline Dictionary-based & 100(53\% NT, 25\% ASD) \newline 35/65 & \textasciitilde  20min per participant & audio \& transcripts & mean 10 years (ASD) \newline 11.29 years (NT) &  Median values for F0 higher and more varied in ASD individuals \newline Identifies 68\% of ASD individuals correctly and 100\% of NT individuals \\
    \hline
    \citet{prudhommeaux2017} & Semantic fluency & 44(n.a.) \newline 22/22 & n.a. & transcripts & 4 - 9 years & No differences in raw item count manually but with similarity measures and machine learning settings \\
    \hline
    \citet{nakai2017} & SVM on single word utterances & 81(29) \newline 51/30 & n.a. & audio & 3 - 10 years & F1: $0.73, 0.56$ \newline Accuracy: $0.76, 0.69$ \\
    \hline
    \citet{hauser2019} & Linear regression model & 140(39) \newline 59/8 1& 6-minute naturalistic conversation samples per participant & audio & middle childhood (8 to 11) \newline adolescence (12 to 17) \newline adulthood (18 and up) & Accuracy (weighted average)): $0.83$ \newline Accuracy: $0.89$\\
    \hline
    \citet{lau2022} & SVM on features from speech rhythm and intonation & English: 94(22) \newline  33/33  \newline Cantonese: 52(16) \newline  24/24 & 20 utterances per participant & audio & English: \newline NT: 12 - 32 \newline  ASD: 6 - 35 \newline Cantonese: \newline NT: 8 - 31 \newline ASD: 8 - 32 & Accuracy rhythm features (English): $0.82$ \newline Accuracy intonation features (English): $0.68$ \newline Accuracy rhythm features (Cantonese): $0.88$ \newline Accuracy intonation features (Cantonese): $0.61$ \newline Accuracy combined features (English and Cantonese): $0.84$\\
    \hline
    \citet{chi2022} & Random Forest (RF) on audio features \newline CNN on spectrograms & 58(23) \newline 38/20 & 77 videos \newline 850 audio clips & audio & median \newline 5 years (ASD) \newline 9.5 years (NT)  & Accuracy (RF): $0.70$ \newline Accuracy (CNN): $0.79$\\
    \hline
    \citet{liu2022} & Transformer-based models & 36(n.a.) \newline 20/16 &  9433 utterances \newline including 3091 ASD & transcripts & 18-30 years (ASD) & Large contextualized language models do not model atypical language very well \\
    \hline
    \citet{plank2023} & Linear L2-regularised L2-loss SVM & 104(66) \newline 69/35 & two 10-minute long conversations for each group of two (including one autistic participant) & audio & mean age 33.15 & Accuracy (balanced): $0.76$ \\
    \hline
    \citet{ashwini2023} & Majority classifier, KNN, Logistic Regression (LR), RF, Gradient Boost and SVM & 76(n.a.) \newline 41/35 & 30-minute free play sessions of 48 children \newline 10-minute free-play task between a child and a parent & transcripts & 3 - 8 years & Accuracy \newline SVM: $0.94$ \newline Majority classifier: $0.56$ \newline KNN: $0.65$ \newline LR: $0.77$ \newline RF: $0.88$ \newline Gradient Boost: $0.77$ \\
    \hline
  \end{tabular}
  \caption{NLP-based approaches to identify specific markers in speech to identify ASD (including Machine learning). "n.a" means that this information was not given, ASD means participants with Autism Spectrum Disorder and NT stands for Neurotypical developing participants.}
  \label{tab:mlapproaches}
  \end{small}
\end{table*}

In line with the medical findings, there are corresponding approaches in natural language processing.
In Table~\ref{tab:mlapproaches} we summarize the various approaches in the NLP community to find markers of ASD in the language.
For each author, we give a short description of the used method.
Column "Support" gives the total amount of participants with additional informatio on the amount of  female participants in brackets.
Additionally, the distribution of NT and ASD participants is given in the second line in the "Support" column.
"Data volume" aims to give a quick overview of the amount of used data in the study.
Please note that not all papers give insight into this (which is described as "n.a." -- not available), whereas others vary greatly in the type of information (Some give the exact length of the used audio data, whereas others mention only the number of videos).
In column "Data type", we clarify whether the experiments were conducted on audio data, transcripts or both.
"Age" gives an overview of  the age of the participants.
It is important to note, that not all the numbers are comparable to each other, as some authors give an age span, some mean values and some median values.
Lastly, column "Results" focuses on a short summary of the results.
This does not only include measurements such as accuracy, but also general observations, e.g. no observed differences in some aspects.

Even though the first entry in the table \cite{bone2012} is not focusing on markers to be used in a NLP setting, it can be used as such (see Section~\ref{ssec:nlpprosodic}).
Although we summarize the participants to be 22 ASD children, it is noteworthy to recognize that the authors differentiate between 17 children with autism and 5 with ASD.

Please note that \citet{parish2016HLT} included 18 non-autistic individuals into their study who have been diagnosed with other medical issues.
%\textcolor{red}{Are these related to the comorbidities?}
These individuals are 94\% male and are assigned to the NT individuals in the table.
The mean age for the non-autistic group with other medical diagnoses is 10.29 years.
As the authors only provide percentage information on the female-to-male ratio of their participants, the specification of the amount of females in the "Support" column differs from the other rows.
Out of the 35 NT participants, 53\% are female, whereas only 25\% of the 65 ASD participants are female.

The numbers of the support in the study of \citet{lau2022} are derived from text and supplemental material (which are conclusive and match) as the numbers do not align with the numbers in Table~1 of their paper.
\citet{liu2022} differentiate in their study between a conversational partner ($n=11$) and experimental participants ($n=9$) for the NT individuals.
We summarized the NT participants in Table~\ref{tab:mlapproaches}.
Please also note that \citet{ashwini2023} derived their data from three different data sets.
As all the data sets provide different measurements, the specification of data volume in our table seems to contradict itself.

In the following sections, we take a closer look into the aforementioned experiments.

\subsection{Prosodic features}
\label{ssec:nlpprosodic}

Prosodic features have also been studied in the NLP domain.

Median values for F0 are both higher and more varied within the ASD and non-ASD mixed clinical group than the NT group (ASD: median: 1.99; non-ASD: median: 1.95; TD: median: 1.47) \cite{parish2016HLT}.
In their study, 75\% of the ASD participants are male, 94\% of the non-ASD mixed clinical participants and 47\% of the NT participants, showing a rather striking imbalance between male and female participants with ASD.

\citet{bone2012} found that descriptions of the voice quality, such as ‘breathy’, ’hoarse’, and ‘nasal’ are common in ASD children.
These quality descriptors can be measured with acoustic features.
\citet{mcallister1998} found shimmer to correlate with breathiness, whereas jitter correlates not only with breathiness but also hoarseness, and roughness (26 male, 24 female participants).
While this is another study with a fairly balanced male-to-female ratio, it is important to note, that \citet{mcallister1998} did not focus on ASD individuals, but on children's voices in general.

\subsubsection{Semantic fluency}
\label{sssec:nlpsemantic}

Semantic fluency is defined as a sub-type of verbal fluency \cite{prudhommeaux2017}.
In semantic fluency tasks, the participants are asked to verbally produce a list of word of a certain category, e.g. animals.
For this task, the participants have a predetermined amount of time, which is usually 60 seconds.

\citet{prudhommeaux2017} analyzed the semantic fluency of responses of autistic individuals (no information on male-female ratio).
According to the authors, there is no standard manual measure of semantic fluency that is able to distinguish autistic children from neurotypical ones.
Apart from manually derived measures, the authors also calculated the mean path similarity for each adjacent word pair in a list of words, the participants generated, by using WordNet.\footnote{\url{https://wordnet.princeton.edu/}}
In order to model multiple dimensions of similarity, the authors also use latent semantic analysis (LSA) and continuous space neural word embeddings as vector-space representations.
The authors use the mean of the set of cosine similarities and also calculate the mean similarity over 100 random permutations of the wordlists generated by the participants in order to gain a "global coherence".
However, there are features derived computationally and the authors find significant differences for autistic and non-autistic groups.
The findings suggest, the subtle differences that are observable via computational measures, such as the ones described above, which could lend support for clinical computational linguistic analysis.

\subsubsection{Disfluencies}
\label{sssec:nlpum}

\citet{parish2016HLT} looked into the production of 'um' in groups of ASD and NT individuals.
To do so, they compared the rate of (\textit{um/(um/uh)}).
In the NT group, 82\% of the filled pauses were produced as \emph{um} by the participants on this study. The ASD participants used \emph{um} as a filled pause only in 61\% of the cases.
The authors observed a minimum value of 58.1\% in the NT group.
More than a third (23 of 65) of the ASD participants fell below that value.

When taking a look at the difference between male an female ASD participants, \citet{parish2016HLT} observe a significant difference in the usage of 'um' and 'uh'.
While male participants filled pauses rather with 'uh' instead of 'um' (56\%), whereas females used 'um' more commonly (75\%).
These findings align with research on typically developing adults in \citet{wieling2016}.

\subsubsection{Speaking rate}
\label{sssec:nlpspeakingrate}

As the studies in Section~\ref{ssec:speak_med} show, speaking rate is an indicator for ASD.
\citet{parish2016HLT} compared the mean word duration in individuals with and without ASD.
The authors found NT individuals to speak the fastest with an overall mean word duration of 376~ms, calculated from 6891 phrases.
The ASD participants reach a much slower speaking rate of 402~ms calculated from 24276 phrases.
Interestingly, the authors had a third group of individuals to compare their results to.
In this group, participants with anxiety, ADHD or sub-threshold ASD symptoms were included.
These participants are in between the NT and ASD group with a mean word duration of 395~ms, calculated from 6640 phrases.

\subsection{Dictionary-based approaches}
\label{ssec:nlpword}

The aforementioned differences in prosodic phrasing were studied in more detail by \citet{parish2016HLT}.
The authors concluded that the word choice as a singular feature works very well to separate NT and ASD individuals.

In their studies, \citet{parish2016HLT} aggregated a list of words that are "ASD-like" and therefore potential indicators of ASD.
Words without a lexical counterpart like imitative or expressive noises, as well as "mhm", "uh" or "eh" are part of this list.
But also seemingly unassuming words like "know", "well", "right", "once", "now", "actually", "first", "year", and "saw" are on this list.
The authors also added "uh" and "w-" which show stuttering-like disfluency.
Additionally, the authors aggregated a list of words, which are "non-ASD" and therefore indicators that the individuals are not autistic.
Part of this list are words like "like", "basketball", "something", "friends", "if", "wrong", "um", or "them".

In their research, the authors use Naive Bayes classification (NB) with leave-one-out cross validation with weighted log-odds-ratios.
They used the informative Dirichlet prior algorithm introduced by \citet{monroe2008}.
By doing so, the authors were able to correctly identify 100\% of the NT participants and 68\% of ASD participants.

\subsection{Machine learning approaches}
\label{ssec:nlpml}

When looking into machine learning approaches, it is important to differentiate between approaches based on the audio data and transcripts thereof.

\subsubsection{Audio Data}
\label{sssec:nlpaudio}

\citet{nakai2017} trained an SVM on single word utterances from 30 ASD individuals (22 male, female 8) and 51 NT individuals (30 male, 21 female), again showing a high imbalance in individuals with ASD.
The authors calculated 24 dimensional features from fundamental frequency (F0) representing pitch.
For this, they extracted static F0 for every 10~ms and calculated the delta F0 from the static F0.
Also, the authors calculated 12 statistics each from the static and delta F0.
Interestingly, the authors compared the performance of their model to the classification of speech therapists.
This lead to a higher F-measure ($0.73$, $0.56$) and accuracy ($0.76$, $0.69$) for the model than the speech therapist.

\citet{hauser2019} trained a linear regression model on 123 features derived from the audio data of 81 ASD individuals (61 male, 20 female) and 59 NT individuals (40 male, 19 female).
The authors computed 12 pause and overlap metrics, 6 segment and turn metrics, 9 speaking rate and word complexity metrics, 80 metrics from the Linguistic Inquiry and Word Count software,\footnote{\url{https://www.liwc.app/}} 5 lexical entropy and diversity measures, and 9 part of speech metrics.
Additionally, the authors computed formality and polarity at conversation level for each speaker by using all words of a speaker in each condition.
The authors down-selected the features by identifying the dimensions with the highest F-value before training the model.
Their model reached a weighted average accuracy of $0.83$ and an accuracy of $0.89$ when taking into account only participants aged 18 to 50.

\citet{lau2022} trained an SVM on features from both speech rhythms as well as intonation for English and Cantonese speech data.
The authors note a severe under-representation of females in their study (38 female, 80 male).
As features to represent the speech rhythm, the authors extracted envelop spectrum (ENV), intrinsic mode functions (IMF), and temporal modulation spectrum (TMS).
This lead to 8640 rhythm-relevant features for the 20 utterances for each of the participants.
For intonation, the authors derived fundamental frequency (F0) for each utterance, which they then concatenated to form a time-normalized F0 contour.
The authors observed rhythm features to be significant in both English (accuracy of $0.82$) and Cantonese (accuracy of $0.88$) classifications, whereas intonation features were only significant for English data (accuracy of $0.68$).
A second experiment, in which the authors did not differentiate between the two languages, rhythm features were found to be significant (accuracy of $0.84$), while intonations features lead to near chance results (accuracy of $0.57$) in correctly predicting ASD.
Interestingly, the features from Cantonese improved the results for the English data ($+0.02$ in accuracy), while it had a negative effect the other way round ($-0.04$ in accuracy).

\citet{chi2022} used data acquired via the \textit{Guess What?} mobile game.
The data included 20 individuals with ASD, which included one female, as well as 38 NT children (22 female).
Also, the median age of ASD children was much lower (5 years) than of the NT children (9.5 years).
The authors trained a Random Forest algorithm on Mel-frequency cepstral coefficients, chroma features, root mean square, spectral centroids, spectral bandwidths, spectral rolloff, and zero-crossing rates.
It reached an accuracy of $0.70$.
Other models such as logistic regression, Gaussian Naive Bayes, and AdaBoosting models did not perform as well as Random Forest.
Additionally, the authors trained a CNN on spectrograms generated via the \textit{Librosa} library for Python,\footnote{\url{https://librosa.org/doc/latest/index.html}} which leads to an accuracy of $0.79$.

\citet{plank2023} trained a linear L2-regularised L2-loss SVM on different features derived from participants aged 18 to 60.
Their data set included 35 ASD individuals (17 males) and 69 NT individuals (21 male).
Of the ASD individuals, two were additionally diagnosed with ADHD.
For their experiment, the authors derived phonetic features using praat.\footnote{\url{https://www.fon.hum.uva.nl/praat/}}
Also, the authors calculated pitch and intensity synchrony, used the uhm-o-meter to extract turns from conversations.
For each turn, they calculated the turn-taking-gap, average pitch, average intensity, and number of syllables in order to calculate the articulation rate.
Also, the authors computed the average of 100 pseudosynchrony or pseudoadaptation values for each synchrony and turn-based adaptation value.
Their SVM reached a balanced accuracy of $0.76$.

\subsubsection{Transcribed Data}
\label{sssec:nlptranscript}

\citet{ashwini2023} trained a majority classifier, K-Nearest Neighbours, Logistic Regression, Random Forest, Gradient Boost, and Support Vector Machine models on the transcripts of ASD and NT children aged 3 to 8 years (no information on male-female ratio was given).
The authors retrieved these transcripts from the Eigsti, Nadig, and Flusberg datasets provided by the Child Language Data Exchange System (CHILDES) databank.
To train their models, the authors used different features: mean length of utterances in words, the number of different word roots, initiative to ask questions, Repetition Prop, child-child discourse coherence, child-partner discourse coherence, Echolalia, Unintell prop, and unexpected words as ASR features (automated stereotypical and repetitive speech).
For syntactic complexity, the authors used, among other features, clause per sentence and  mean length of sentence.
Additionally, POS tag features and the corresponding frequencies are used.
These feature sets were also combined in different variations.
The best results were obtained by combining all features and using SVMs (Accuracy of $0.94$).

\subsection{Transformer-based models}
\label{ssec:nlptransformer}

Interestingly, there is very little transformer-based research or research based on deep learning in general.

\citet{liu2022} built different transformer-based models in order to identify linguistic features for autistic language.
To identify features that are associated with social aspects of communication, the authors used a corpus of conversations between adults with and without ASD (no information on male-female ratio was given).
These conversations have been recorded while the participants were engaging in collaborative tasks, which were meant to resemble workplace activities.
However, it is important to note, that the experiments are only conducted on written data in form of transcriptions but not the speech data itself.
The model performed much worse for ASD participants than for NTs.
The authors concluded that individuals with ASD use a more diverse set of strategies for some of the social linguistic functions.
In general, the results of \citet{liu2022} show, that large contextualized language models do not model atypical language very well.
A reason for that might be the bias that arises from trained models mostly on news and web data.
It is not surprising to the authors that models trained primarily on this data do not perform very well on ASD language.

\section{Discussion \& Conclusion}
\label{sec:conclusion}

This survey takes a look at research in the detection of identifier of autism in speech.
While there is already some research, there are still some observable shortcomings.

Firstly, the mentioned studies show a massive under-representation of females in autism studies in general.
While there are mostly at least some females participating the the mentioned studies, in comparison to their male counterparts, they make up fewer of the participants.
This shows even more so in NLP approaches, which might be because of the lack of data gathered from females on the spectrum.

Secondly, it is noticeable that most NLP experiments use traditional machine learning approaches like SVMs, Naive Bayes or Linear Regession.
Interestingly, there are very few experiments conducted with transformers or deep learning methods in general.
Further research should therefore investigate, whether transformers or other deep learning methods might be a good fit for the classification of ASD and possibly even improve the results we see so far.
If this is not the case, it might be possible, that simpler algorithms fit the task better, which should also be addressed in further research.
However, the lack of data might be another possible reason for the focus on more traditional methods as they require less training data.

Thirdly, while there is some research on the transcripts, compared to the amount of experiments performed on audio data, there is considerably less research on transcribed data.
It should be investigated if NLP approaches on the transcripts can reach good results in detecting identifiers of autism and whether it can be further improved.

Lastly, we could not find any research combining the features from both the transcriptions and the audio input.
Future work should therefore investigate, if features from either could improve the results of the other or if they maybe even hinder each other in getting good results.

\section*{Limitations}

While trying to include data from as many different backgrounds as possible, this survey is not able to include all existing cultural or ethnical groups.
Also, our main focus lays on adults on the spectrum, however, we also included studies with children of various ages.
Nevertheless, as some studies show, the differences in age lead to very different outcomes.
For this reason it was not possible for us to include all possible age groups and variations thereof.
Therefore, it is not possible to generalise the findings of this paper to all individuals on the spectrum.

\section*{Ethics Statement}
Even though we look into identifiers for ASD in voice, speech and language, it is important to note, that we do not intend to say that these findings can be used to automatically classify the disorder.
Our findings should therefore not be used in any way to replace a professional diagnosis, but rather the described indicators of ASD might be of use to support a diagnosis.
We are not responsible for how the data cited in this survey has been collected and/or annotated.